\documentclass[11pt]{article}

\usepackage[final]{acl}

\usepackage{times}
\usepackage{latexsym}

\usepackage[T1]{fontenc}

\usepackage[utf8]{inputenc}

\usepackage{microtype}

\usepackage{inconsolata}

\usepackage{graphicx}
\usepackage{hyperref}
\usepackage{url}

\usepackage{pifont}
\usepackage{xspace}
\usepackage{enumitem}
\usepackage{amsthm}
\usepackage{graphicx}
\usepackage{amssymb}
\usepackage{multirow}
\usepackage{booktabs}
\usepackage{scalerel}
\usepackage{makecell}
\usepackage{wrapfig}
\usepackage{colortbl}
\usepackage{xcolor} 
\usepackage{soul} 
\usepackage{tcolorbox}
\usepackage[normalem]{ulem}
\usepackage{microtype} 
\usepackage{caption}
\usepackage{makecell}
\usepackage{array}
\usepackage{booktabs}
\usepackage[table]{xcolor}
\usepackage{graphicx}
\usepackage{booktabs,tabularx,array}
\newcolumntype{L}[1]{>{\raggedright\arraybackslash}p{#1}} 
\usepackage{bbm}
\usepackage{booktabs}
\usepackage{multirow}
\usepackage{amssymb}   
\usepackage{graphicx}   
\usepackage{siunitx}
\usepackage{amsmath}

\newcommand{\ie}{\textit{i.e., }}
\newcommand{\eg}{\textit{e.g., }}

\usepackage{xcolor}
\usepackage{makecell}
\usepackage{booktabs}
\usepackage{graphicx}

\newcommand{\orig}[1]{\textcolor{gray!90}{#1}}
\newcommand{\fdlt}[1]{\textcolor{red!70}{#1}} 

\title{AutoRubric: Rubric-Based Generative Rewards \\ for Faithful Multimodal Reasoning}

\author{
Mengzhao Jia$^{1}$, Zhihan Zhang$^{1}$, Ignacio Cases$^{2}$, Zheyuan Liu$^{1}$, \\
\textbf{Meng Jiang}$^{1}$, \textbf{Peng Qi}$^{2}$ \\
$^{1}$University of Notre Dame, 
$^{2}$Uniphore \\
\texttt{mjia2@nd.edu}, \texttt{peng.qi@uniphore.com} \\
}

\begin{document}
\maketitle
\begin{abstract}
Multimodal large language models (MLLMs) have rapidly advanced from perception tasks to complex multi-step reasoning, yet reinforcement learning with verifiable rewards (RLVR) often leads to spurious reasoning since only the final-answer correctness is rewarded. To address this limitation, we propose AutoRubric, a framework that integrates RLVR with process-level supervision through automatically collected rubric-based generative rewards. Our key innovation lies in a scalable self-aggregation method that distills consistent reasoning checkpoints from successful trajectories, enabling problem-specific rubric construction without human annotation or stronger teacher models. By jointly leveraging rubric-based and outcome rewards, AutoRubric achieves state-of-the-art performance on six multimodal reasoning benchmarks and substantially improves reasoning faithfulness in dedicated evaluations.

\end{abstract}
\section{Introduction}
\begin{figure*}[t]
    \centering
    \includegraphics[width=0.95\textwidth]{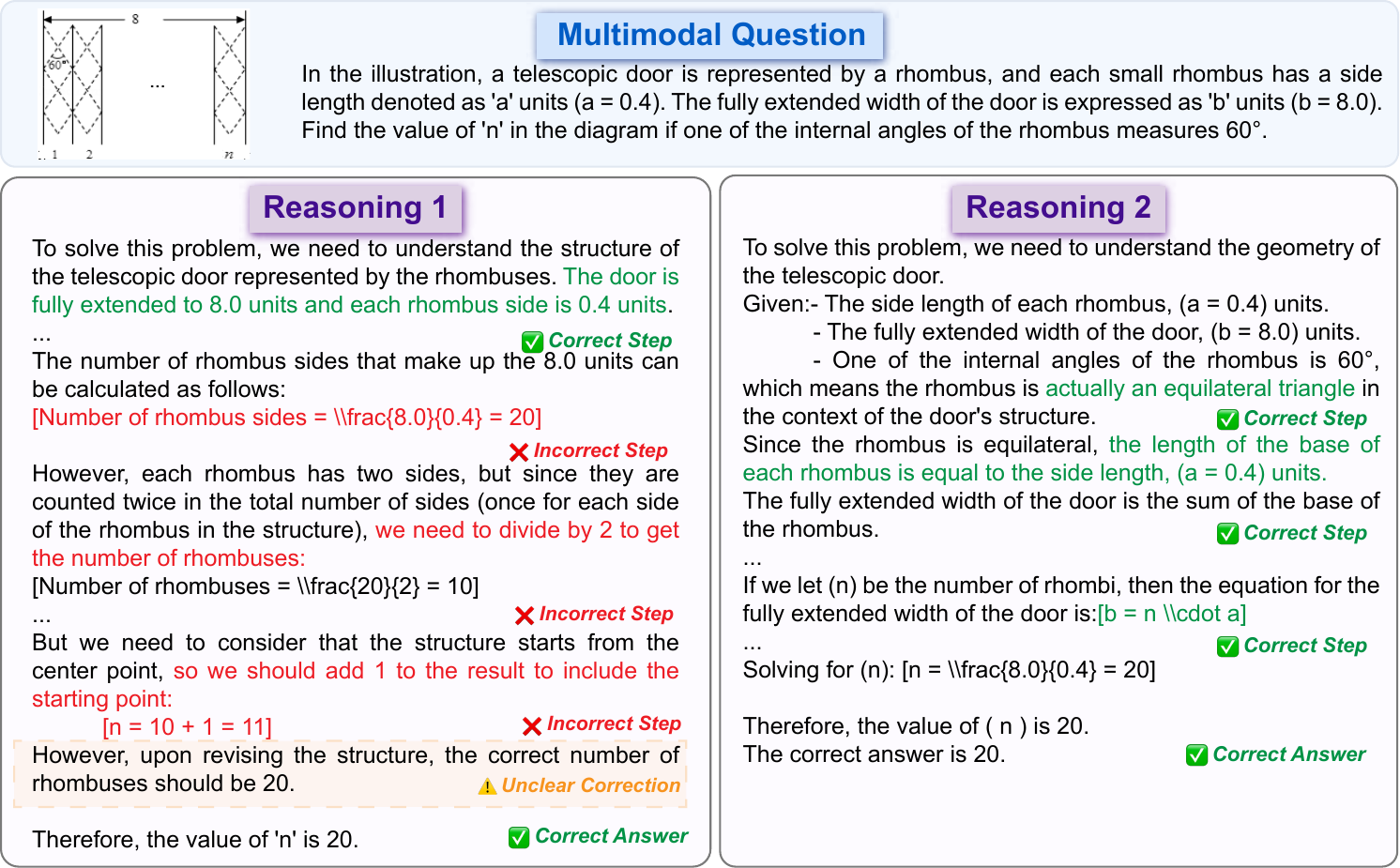}
    \caption{Illustration of a multimodal reasoning question together with two model-generated reasoning traces that both reach the correct answer. \textit{Reasoning 1} contains logical flaws—such as incorrectly halving rhombus sides and inconsistently switching from 11 to 20 without reconciliation—while \textit{Reasoning 2} proceeds with fully consistent step-by-step logic. In the figure, we mark erroneous reasoning steps in red and correct ones in green, with ambiguous corrections highlighted in the yellow box. Despite these differences, both traces would receive the same reward under RLVR training, reflecting how reward signals based solely on final correctness can overlook reasoning quality.}
    \label{fig:intro}
    \vspace{-1.7em}
\end{figure*}

Multimodal Large Language Models (MLLMs) have rapidly progressed from simple perception tasks such as visual question answering and image captioning to complex multi-step reasoning tasks~\citep{Mulberry,OThink-MR1,Skywork-R1V}. Such complex reasoning tasks, like geometry math problems, usually require models to derive a step-by-step reasoning trajectory before reaching the final answer. Reinforcement learning with verifiable rewards (RLVR), which assigns training rewards only according to the correctness of the final answer, is a popular method in optimizing MLLMs on reasoning tasks due to its simplicity and efficiency~\citep{Mm-eureka,NoisyRollout,Mixed-R1}. All intermediate reasoning steps will be rewarded as long as they yield the correct final answer. Unfortunately, it is prevalent for the model to learn spurious reasoning under such a rewarding paradigm: models may exploit shortcuts or generate contradictory intermediate steps that still land on the right output, effectively “hacking” the training objective. As illustrated in Figure~\ref{fig:intro}, two distinct trajectories can both reach the correct answer, but one does so by introducing flawed logic and abruptly altering results, while the other follows a coherent, step-by-step derivation. Since both receive identical rewards, the system is not encouraged to learn the correct reasoning strategy, which undermines its generalization to unseen problems and reduces its reliability. Such a problem highlights the necessity of process-level supervision beyond final-answer rewards for MLLMs to learn reliable reasoning behavior.

To incorporate process-level supervision into reasoning training, a common approach is to leverage pre-trained progress reward models (PRMs), which score intermediate reasoning steps based on their correctness~\citep{VisualPRM,URSA-PRM}. While PRMs provide fine-grained supervision, they are often vulnerable to distribution shifts, which can lead to unreliable reward estimates when applied to problems from different domains or reasoning steps generated by unseen policy models~\citep{prm-Distribution-Shift}. Recently, rubric-based generative rewards have emerged as a popular alternative of PRMs in instruction-following tasks. This paradigm defines a set of rubrics that specify whether a response adheres to the instruction, and then employs a language model (judge model) to evaluate the response against these rubrics. Compared to traditional reward models, rubric-based approaches offer more robust and interpretable reward signals~\citep{Checklists-Are-Better-Than-Reward-Models,Rubric-Anchors}. However, while rubrics in instruction-following tasks can often be directly derived from the input instruction, extending this paradigm to multimodal reasoning tasks is non-trivial, as the ground-truth reasoning trajectory is usually unknown. Consequently, designing reliable rubrics and effectively integrating them into RLVR for multimodal reasoning remains an open challenge.

Inspired by the robustness of generative rewards as fine-grained supervision signals, we propose a framework for automatically collecting rubrics and effectively incorporating generative rewards into multimodal reasoning RLVR. Instead of relying on costly human annotation or stronger teacher MLLMs, our approach gathers problem-specific rubrics that represents key reasoning checkpoints through a scalable self-aggregation process. Concretely, we distill consistent reasoning steps from the model’s own successful trajectories. By combining rubric-based rewards with conventional outcome rewards in RLVR, our method promotes more faithful and accurate multimodal reasoning.

With this framework, we train a model named \textbf{AutoRubric}, which demonstrates superior performance as well as faithfulness. Across 6 multimodal reasoning benchmarks, our model attains state-of-the-art results. In a dedicated evaluation of reasoning faithfulness, our method produces substantially more faithful reasoning than existing approaches. Ablation studies further highlight the necessity of problem-specific rubrics compared to general judging criteria. Moreover, detailed analysis of the training dynamics shows that our framework effectively stabilizes training. To facilitate further research, we will release the constructed rubric dataset and code.

\section{Related Work }
\vspace{-0.7em}
\paragraph{Reinforcement Learning in MLLM Reasoning.}
Multimodal large language models (MLLMs) have rapidly progressed by integrating visual encoders with large language models for cross-modal understanding and reasoning. Early advances mainly relied on multimodal supervised finetuning with large-scale instruction data, such as InstructBLIP~\citep{InstructBLIP} and LLaVA~\citep{LLaVA}. More recently, reinforcement learning with verifiable rewards (RLVR)~\citep{GRPO} has emerged as a key paradigm for improving multimodal reasoning, using rule-based verification of final answers for policy optimization. Prior work largely follows two directions: (1) strengthening reasoning capability before RL by distilling multimodal chain-of-thought data from teacher models, e.g., Vision-R1~\citep{vision-r1} and~\cite{cold-start-mmreasoning}; and (2) enriching supervision beyond answer correctness, such as annotated key steps in R1-VL~\citep{R1-VL}, visual perception rewards in Vision-SR1~\citep{Perception-R1}, or reflection-based rewards in SRPO~\citep{srpo}. However, these methods often emphasize isolated aspects of reasoning and rely heavily on costly proprietary MLLM annotations.

In contrast, AutoRubric derives problem-specific rubrics directly from multiple successful reasoning trajectories without proprietary supervision. By aggregating consistent reasoning steps and filtering spurious ones, AutoRubric provides effective process-level rewards that improve reasoning fidelity and discourage shortcut solutions.

\vspace{-0.5em}
\paragraph{Rubrics in RL.}
Since some instructions are not compatible with RLVR, and considering that traditional reward models often struggle to generalize to out-of-distribution inputs, recent literature has proposed the use of explicit rubrics to assign rewards for RL~\cite{Rubrics-as-Rewards,Rubric-Anchors}. These rubrics can be either query-agnostic, focusing on general response quality, or query-specific, tailored to the nuanced requirements of a given prompt~\cite{advancedif}. Typically, these rubrics are integrated into a prompt for an LLM-as-a-judge to evaluate the policy model's response. Existing methods for rubric generation include manual annotation~\cite{advancedif}, derivation from teacher LLM responses~\cite{checklists-better-than-RM,VerIF}, or contrastive analysis of responses with varying quality~\cite{OpenRubrics}. While prior work mainly applies rubrics to general instruction-following tasks where verifiable rewards are not available, our work demonstrates that rubrics can be synergized with verifiable rewards in multimodal reasoning to enhance the accuracy and faithfulness of intermediate reasoning chains.

\begin{figure*}[t]
    \centering
    \includegraphics[width=0.95\textwidth]{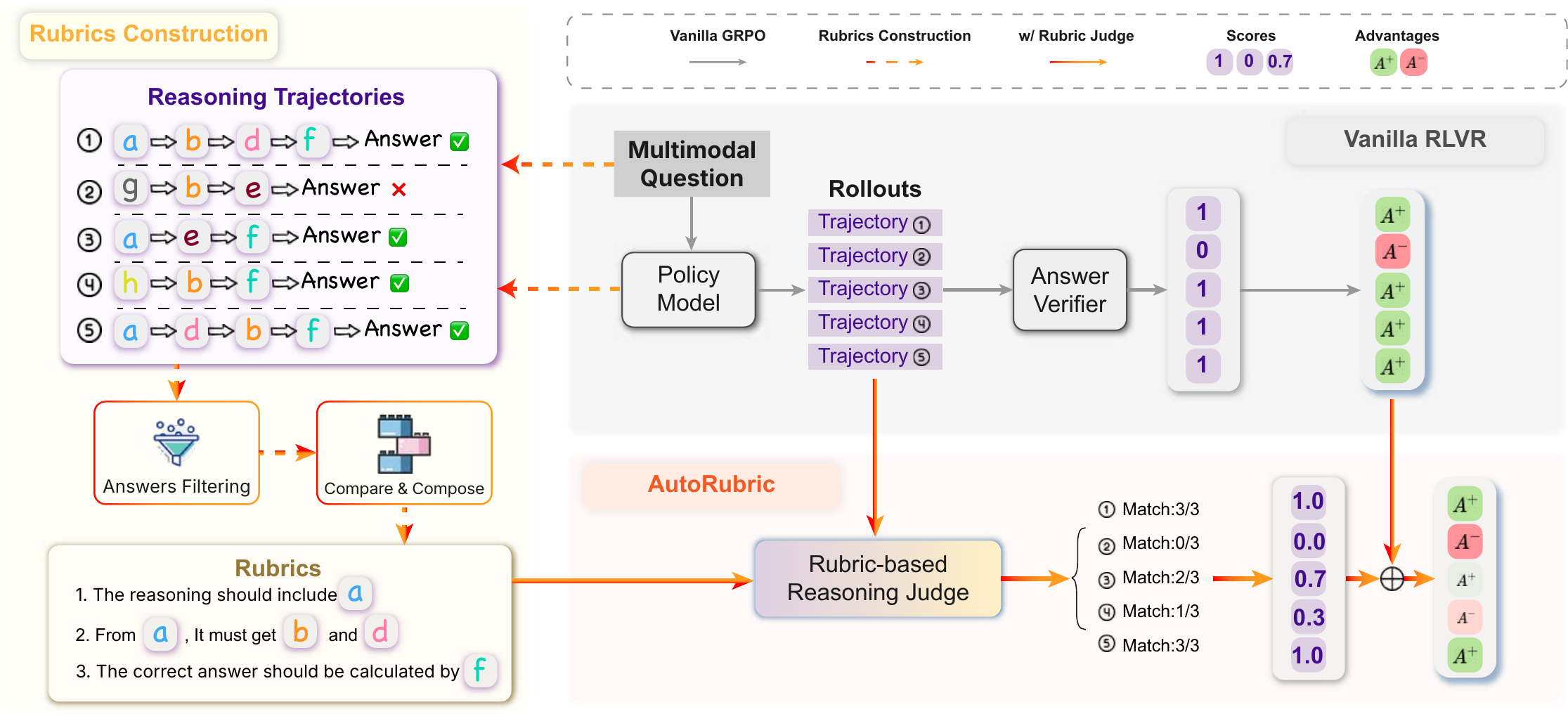}
    \caption{Our framework augments RLVR with rubric-based reasoning rewards. \textbf{Left}: Rubrics are automatically constructed by aggregating \textit{common steps} across multiple correct reasoning trajectories, yielding problem-specific rubrics criteria without human annotation. \textbf{Right}: While vanilla RLVR cannot distinguish reasoning quality among trajectories with the same answer, rubric-based scoring enables step-level differentiation and provides richer training signals for faithful-oriented reasoning.}
    \label{fig:method}
    \vspace{-0.6cm}
\end{figure*}

\section{Method}
Our method aims to enhance the reasoning capabilities of vision-language models through a reinforcement learning framework. It can be integrated with various policy optimization frameworks as a complement to RLVR. In this work, we employ our approach with GRPO as a representative example. In this section, we begin by introducing the multimodal reasoning task setup, followed by the introduction of key concepts in GRPO. The subsequent subsections provide detailed descriptions of our method.

\subsection{Preliminary}
\paragraph{Problem Formulation}
In this work, we focus on using MLLMs for solving multimodal reasoning task. 
Assume we are given a collection of $N$ multimodal reasoning instances denoted as $\mathcal{D} = \{x_i\}_{i=1}^N$. 
Each instance $x_i \in \mathcal{D}$ contains a visual input $\mathcal{V}_i$, a textual query $\mathcal{Q}_i$, and its labeled answer $a_i$. 
Our objective is to train a policy model that learns a function $\mathcal{F}: (\mathcal{V}_i, \mathcal{Q}_i) \mapsto a_i$.

To improve reasoning performance, the model is instructed to generate a 
token sequence that contains intermediate reasoning steps with the final answer:
\[
s_{i,t} \sim \pi_\theta\!\left(\,\cdot \mid \mathcal{V}_i, \mathcal{Q}_i, s_{i,<t}\,\right),
\quad t = 1,\ldots,T_i,
\]
where the trace $s_{i,1:T_i}$ jointly encodes the reasoning process and ends with the final answer $\hat{a}_i$.

\paragraph{Group Relative Policy Optimization.}
We adopt Group Relative Policy Optimization (GRPO) to optimize the policy model.
GRPO is a variant of PPO that removes the need for a separately trained value function,
and instead relies on relative comparisons among multiple responses sampled for the same query.
This design yields a lightweight and stable training procedure.

Given a query $q$, we sample a group of $G$ responses $\{o_i\}_{i=1}^G$ from the old policy
$\pi_{\theta_{\text{old}}}$.
Each response $o_i = (o_{i,1}, \dots, o_{i,|o_i|})$ receives a scalar reward $r_i$.
GRPO computes a group-normalized advantage $\hat{A}_i$ from $\{r_i\}_{i=1}^G$,
which serves as the relative learning signal shared across all tokens of $o_i$.

The policy $\pi_\theta$ is then updated using a clipped importance sampling objective
with KL regularization toward a fixed reference policy $\pi_{\text{ref}}$.
We denote the token-level importance ratio as
$\rho_{i,t}(\theta) = \pi_\theta(o_{i,t} \mid q, o_{i,<t}) /
\pi_{\theta_{\text{old}}}(o_{i,t} \mid q, o_{i,<t})$.
Full optimization details are provided in Appendix~\ref{app:grpo}.

\subsection{Integrating LLM-As-a-Judge into RLVR}
While RLVR optimizes the policy model solely based on answer correctness, this signal is often sparse and insufficient to capture the quality of intermediate reasoning. To provide a richer supervision signal, we incorporate an additional \emph{rubric-based reasoning reward} derived from a language model acting as a trajectory judge inspired.

\paragraph{Rubric-guided Scoring.}
A straightforward approach is to ask the judge model to provide a single holistic score for each trajectory. Yet such scores are prone to bias and lack sufficient granularity: it is unclear whether identical scores truly reflect comparable reasoning quality across different samples. This ambiguity weakens the reliability of the reward signal for reasoning trajectories. To mitigate these issues, we guide the reasoning reward process with problem-specific rubrics $\mathcal{C}^x = \{c_1, \dots, c_m\}$. Each rubric item $c_j$ specifies a key reasoning checkpoint that is expected to appear in a logically sound trajectory. Given a candidate trajectory $\tau$, the judge model verifies whether $\tau$ satisfies each checkpoint. Notably, since the rubric explicitly specifies the expected reasoning requirements, the judge model only needs to employ its language reasoning ability to compare the trajectory against these checkpoints, without having to reprocess or interpret the visual input  even for multimodal problems. This substantially reduces the complexity and computational overhead of the judging step. Let $\mathbbm{1}[\tau \vDash c_j]$ denote an indicator function that equals $1$ if $c_j$ is satisfied, and $0$ otherwise. The rubric-based reasoning reward is then computed as the fraction of satisfied checkpoints:
\begin{equation}
 r^{\text{rubric}}_i = \frac{1}{|\mathcal{C}^x|} \sum_{j=1}^{|\mathcal{C}^x|} \mathbf{1}[\tau \vDash c_j].
\end{equation}

\paragraph{Combining outcome and rubric-based rewards.}
The rubric-based reward $r^{\text{rubric}}$ is integrated with the conventional 
outcome reward $r^{\text{ans}}$ that indicates whether the final prediction $\hat{a}$ 
matches the ground truth with a weighted combination:
\begin{equation}
r_i = \lambda r^{\text{ans}}_i + (1-\lambda) r^{\text{rubric}}_i,
\end{equation}
where $\lambda \in [0,1]$ controls the impact of the rubric-based reward.
During policy optimization, the combined reward $r_i$ is assigned to each sampled 
trajectory, and the normalized group-relative advantages are computed following 
the GRPO framework. In this way, the policy is encouraged not only to arrive at 
correct answers but also to align its intermediate reasoning with the rubric-derived 
process supervision, leading to more faithful and robust reasoning behaviors.

\subsection{Aggregation-based Rubric Generation}

Existing approaches to acquire process supervision signals often resort to compare with manually annotated or stronger proprietary MLLMs' reasoning trajectories. Manual annotation is prohibitively expensive. Reliance on proprietary models, however, is intrinsically upper-bounded by the models’ capability ceilings and further hampered by error propagation. Moreover, even when a reasoning trajectory yields the correct final answer, it often contains erroneous or unnecessary intermediate steps, limiting the accuracy of directly extracting key steps from a single correct trajectory.

\begin{table}[t]
  \centering
  \small
  \caption{Summary statistics of the rubric sets of the training samples.}
  \vspace{-0.5em}
  \begin{tabular}{@{}lr@{}}
    \toprule
    \multicolumn{2}{@{}l@{}}{\textbf{Overview}}\\
    \midrule
    \# Training Samples & 38,870 \\
    \# Rubric sets      & 26,144 \\
    Coverage           & 67.26\% \\
    Avg. / Total words & 80.65 / 2,107,756 \\
    \addlinespace
    \multicolumn{2}{@{}l@{}}{\textbf{Rubric Criteria Statistics}}\\
    \midrule
    Avg. criterion     & 3.47 \\
    Avg. / Max words   & 23.25 / 198 \\
    \bottomrule
  \end{tabular}
  \vspace{-1em}
  \label{tab:rubric_stats}
\end{table}

To mitigate this issue, we take inspiration from the idea of \emph{test-time scaling}~\citep{Self-Consistency,Large-Language-Monkeys}, which suggests that increasing inference computation, \eg sampling multiple reasoning attempts, increases the likelihood that the majority will converge to a correct solution. 
Analogously, we propose to \emph{aggregate step-level consistency across the model’s own successful trajectories}. 
The key intuition is that if a particular step consistently appears in many correct trajectories, it is likely to represent a causally essential component of the reasoning process; in contrast, steps that appear only sporadically are more likely to be spurious or unnecessary. Figure~\ref{fig:method} demonstrate this process: 4 reasoning trajectories reach the correct answer, but their intermediate steps are not identical. By comparing steps, we can see some steps consistently recur across multiple correct trajectories (\eg Reasoning from step $a$ to derive $b,d$, and calculating final answer with step $f$). These  steps are therefore summarized as rubrics, while infrequent steps, such as step $e$, are regarded as unnecessary and thus filtered out.

Given a multimodal reasoning problem $x$, we first sample $K$ reasoning trajectories $\{\tau^{(k)}\}_{k=1}^{K}$ from the current policy. 
Among them, we retain the subset $\mathcal{S} \subseteq \{\tau^{(k)}\}$ whose final answers match the verifiable ground truth. 
We then prompt an LLM to compare trajectories in $\mathcal{S}$ and summarize their common steps into an ordered set of key checkpoints:
\[
\mathcal{C}^x = \{ c_1, c_2, \dots, c_m \},
\]
where each $c_i$ denotes a reasoning checkpoint distilled from recurring steps across correct trajectories. 
These checkpoints are organized into $\mathcal{C}^x$, a structured collection of checkpoints that encodes the essential reasoning requirements for derive the correct answer, which further serve as the problem-specific rubrics for the LLM-as-a-Judge reasoning rewarding during training.

\section{Experiments}

\begin{table*}[t]
  \centering
  \caption{
  Performance comparison of open vision-language reasoning models on multimodal reasoning benchmarks.
  In each cell, we report
  \scalebox{0.6}{%
    [\,\shortstack[c]{{Strict Accuracy}\\\orig{Standard Accuracy} / \fdlt{False Positive}}\,]
  },
  where Strict Accuracy and False Positive terms are defined in Sec.\ref{sec:stricacc}.
  Larger Strict Accuracy and Smaller absolute False Positive indicates more faithful reasoning.
  The best results are highlighted in \textbf{bold}, while the second-best are \underline{underlined}.
  }
  \resizebox{\textwidth}{!}{
  \begin{tabular}{lcccccc}
    \toprule
    \textbf{Models}
    & \textbf{Avg.}
    & \textbf{MathVision}
    & \textbf{MathVista}
    & \textbf{MMMU}
    & \textbf{MMMU Pro}
    & \textbf{Wemath} \\
    \midrule

   Qwen2.5-VL-7B~\citep{qwen2.5-vl}
    & \makecell[c]{45.24 \\ {\footnotesize \orig{49.04} / \fdlt{-3.80}}}
    & \makecell[c]{21.60 \\ {\footnotesize \orig{26.20} / \fdlt{-4.60}}}
    & \makecell[c]{66.00 \\ {\footnotesize \orig{68.60} / \fdlt{-2.60}}}
    & \makecell[c]{49.80 \\ {\footnotesize \orig{55.00} / \fdlt{-5.20}}}
    & \makecell[c]{33.80 \\ {\footnotesize \orig{37.40} / \fdlt{-3.60}}}
    & \makecell[c]{55.00 \\ {\footnotesize \orig{58.00} / \fdlt{-3.00}}} \\

MM-Eureka~\citep{Mm-eureka}
    & \makecell[c]{47.83 \\ {\footnotesize \orig{50.57} / \fdlt{-2.74}}}
    & \makecell[c]{24.08 \\ {\footnotesize \orig{27.47} / \fdlt{-3.39}}}
    & \makecell[c]{69.80 \\ {\footnotesize \orig{71.80} / \fdlt{-2.00}}}
    & \makecell[c]{51.00 \\ {\footnotesize \orig{52.78} / \fdlt{-1.78}}}
    & \makecell[c]{35.66 \\ {\footnotesize \orig{36.47} / \fdlt{-0.81}}}
    & \makecell[c]{58.62 \\ {\footnotesize \orig{64.31} / \fdlt{-5.69}}} \\

R1-VL~\citep{R1-VL}
    & \makecell[c]{38.44 \\ {\footnotesize \orig{40.89} / \fdlt{-2.45}}}
    & \makecell[c]{20.43 \\ {\footnotesize \orig{23.39} / \fdlt{-2.96}}}
    & \makecell[c]{53.00 \\ {\footnotesize \orig{54.90} / \fdlt{-1.90}}}
    & \makecell[c]{42.00 \\ {\footnotesize \orig{46.56} / \fdlt{-4.56}}}
    & \makecell[c]{26.18 \\ {\footnotesize \orig{27.75} / \fdlt{-1.57}}}
    & \makecell[c]{50.57 \\ {\footnotesize \orig{51.84} / \fdlt{-1.27}}} \\

NoisyRollout~\citep{NoisyRollout}
    & \makecell[c]{\underline{50.66} \\ {\footnotesize \orig{52.39} / \fdlt{-1.73}}}
    & \makecell[c]{\underline{26.97} \\ {\footnotesize \orig{28.29} / \fdlt{-1.32}}}
    & \makecell[c]{\underline{71.50} \\ {\footnotesize \orig{73.00} / \fdlt{-1.50}}}
    & \makecell[c]{\underline{53.22} \\ {\footnotesize \orig{56.11} / \fdlt{-2.89}}}
    & \makecell[c]{37.34 \\ {\footnotesize \orig{38.44} / \fdlt{-1.10}}}
    & \makecell[c]{\underline{64.25} \\ {\footnotesize \orig{66.09} / \fdlt{-1.84}}} \\

VLAA-Thinker~\citep{vlaa-thinker}
    & \makecell[c]{45.62 \\ {\footnotesize \orig{49.16} / \fdlt{-3.54}}}
    & \makecell[c]{23.55 \\ {\footnotesize \orig{26.88} / \fdlt{-3.33}}}
    & \makecell[c]{67.70 \\ {\footnotesize \orig{70.10} / \fdlt{-2.40}}}
    & \makecell[c]{48.00 \\ {\footnotesize \orig{52.33} / \fdlt{-4.33}}}
    & \makecell[c]{33.47 \\ {\footnotesize \orig{36.42} / \fdlt{-2.95}}}
    & \makecell[c]{55.40 \\ {\footnotesize \orig{60.06} / \fdlt{-4.66}}} \\

Perception-R1~\citep{Perception-R1}
    & \makecell[c]{50.09 \\ {\footnotesize \orig{51.24} / \fdlt{-1.15}}}
    & \makecell[c]{25.89 \\ {\footnotesize \orig{26.84} / \fdlt{-0.95}}}
    & \makecell[c]{71.40 \\ {\footnotesize \orig{72.00} / \fdlt{-0.60}}}
    & \makecell[c]{50.89 \\ {\footnotesize \orig{52.89} / \fdlt{-2.00}}}
    & \makecell[c]{\underline{38.38} \\ {\footnotesize \orig{39.13} / \fdlt{-0.75}}}
    & \makecell[c]{63.91 \\ {\footnotesize \orig{65.34} / \fdlt{-1.43}}} \\

ThinkLite-VL~\citep{thinklite-vl}
    & \makecell[c]{47.84 \\ {\footnotesize \orig{51.19} / \fdlt{-3.35}}}
    & \makecell[c]{22.53 \\ {\footnotesize \orig{24.54} / \fdlt{-2.01}}}
    & \makecell[c]{68.70 \\ {\footnotesize \orig{73.30} / \fdlt{-4.60}}}
    & \makecell[c]{50.56 \\ {\footnotesize \orig{53.67} / \fdlt{-3.11}}}
    & \makecell[c]{36.82 \\ {\footnotesize \orig{39.42} / \fdlt{-2.60}}}
    & \makecell[c]{60.57 \\ {\footnotesize \orig{65.00} / \fdlt{-4.43}}} \\

Vision-G1~\citep{vision-g1}
    & \makecell[c]{48.37 \\ {\footnotesize \orig{53.92} / \fdlt{-5.55}}}
    & \makecell[c]{25.82 \\ {\footnotesize \orig{28.75} / \fdlt{-2.93}}}
    & \makecell[c]{70.00 \\ {\footnotesize \orig{\textbf{76.40}} / \fdlt{-6.40}}}
    & \makecell[c]{47.67 \\ {\footnotesize \orig{53.78} / \fdlt{-6.11}}}
    & \makecell[c]{34.45 \\ {\footnotesize \orig{\underline{39.48}} / \fdlt{-5.03}}}
    & \makecell[c]{63.91 \\ {\footnotesize \orig{\underline{71.21}} / \fdlt{-7.30}}} \\

VL-Rethinker~\citep{VL-Rethinker}
    & \makecell[c]{49.22 \\ {\footnotesize \orig{\underline{54.15}} / \fdlt{-4.93}}}
    & \makecell[c]{25.69 \\ {\footnotesize \orig{\textbf{31.12}} / \fdlt{-5.43}}}
    & \makecell[c]{70.80 \\ {\footnotesize \orig{73.90} / \fdlt{-3.10}}}
    & \makecell[c]{52.22 \\ {\footnotesize \orig{\textbf{57.11}} / \fdlt{-4.89}}}
    & \makecell[c]{35.95 \\ {\footnotesize \orig{39.42} / \fdlt{-3.47}}}
    & \makecell[c]{61.44 \\ {\footnotesize \orig{69.20} / \fdlt{-7.76}}} \\

VL-Reasoner~\citep{VL-Rethinker}
    & \makecell[c]{48.56 \\ {\footnotesize \orig{53.38} / \fdlt{-4.82}}}
    & \makecell[c]{25.23 \\ {\footnotesize \orig{29.87} / \fdlt{-4.64}}}
    & \makecell[c]{70.60 \\ {\footnotesize \orig{{74.80}} / \fdlt{-4.20}}}
    & \makecell[c]{52.00 \\ {\footnotesize \orig{56.22} / \fdlt{-4.22}}}
    & \makecell[c]{35.66 \\ {\footnotesize \orig{38.96} / \fdlt{-3.30}}}
    & \makecell[c]{59.31 \\ {\footnotesize \orig{67.07} / \fdlt{-7.76}}} \\

\textbf{AutoRubric}
    & \makecell[c]{\textbf{53.24} \\ {\footnotesize \orig{\textbf{55.26}} / \fdlt{{-2.02}}}}
    & \makecell[c]{\textbf{29.14} \\ {\footnotesize \orig{\underline{30.49}} / \fdlt{{-1.35}}}}
    & \makecell[c]{\textbf{73.60} \\ {\footnotesize \orig{\underline{75.80}} / \fdlt{{-2.20}}}}
    & \makecell[c]{\textbf{54.67} \\ {\footnotesize \orig{\underline{56.56}} / \fdlt{{-1.89}}}}
    & \makecell[c]{\textbf{39.60} \\ {\footnotesize \orig{\textbf{40.98}} / \fdlt{{-1.38}}}}
    & \makecell[c]{\textbf{69.20} \\ {\footnotesize \orig{\textbf{72.47}} / \fdlt{{-3.27}}}} \\

    \bottomrule
  \end{tabular}
  }
    \vspace{-1em}
  \label{tab:main-res}
\end{table*}

\subsection{Experimental Setup}
\paragraph{Implementation Details.}
In our experiments, we use Qwen2.5-VL-7B-IT~\citep{qwen2.5-vl} as the base model and train it with the verl\footnote{\url{https://github.com/volcengine/verl}.} framework. We adopt GRPO~\cite{GRPO} as the policy update algorithm during training. We train the model with \textit{ViRL-39K} dataset proposed by~\citet{VL-Rethinker} for 4 epochs with a constant learning rate of 1e-6. We adopt 512 as the rollout batch size and 128 as the global policy update batch size. We set the rollout number to 8 with a sampling temperature of 1.0. For rubric-based reasoning rewards, we employ an open-sourced LLM as the judge model\footnote{\url{https://huggingface.co/openai/gpt-oss-20b}.}. The KL coefficient is fixed at 0.01. All experiments are run on a single node equipped with 8 H100 GPUs. The full set of prompts used in rubric construction and rubric-based rollouts scoring in training, is provided in the Appendix.

\paragraph{Benchmarks.}
We evaluate model performance along two dimensions. For general multimodal reasoning, we adopt MMMU~\citep{mmmu} and MMMU-Pro~\citep{mmmu-pro}, which cover diverse subjects on multimodal reasoning. For multimodal mathematical reasoning, we include three challenging benchmarks: MathVista~\citep{mathvista}, MATH-Vision~\citep{mathvision}, and WeMATH~\citep{wemath}, each designed to test different aspects of multimodal mathematical problem-solving skills. 

\paragraph{Evaluation Metrics.}
\label{sec:stricacc}
To comprehensively assess both answer correctness and reasoning reliability, we adopt two distinct accuracy evaluation metrics, introduced below.

\textbf{Standard Accuracy} evaluates a model solely based on whether its final predicted answer matches the ground-truth answer.
Formally, for each prediction $i \in \mathcal{D}$, a binary indicator $c_i \in \{0,1\}$ is assigned, where $c_i = 1$ if the final answer is correct and $c_i = 0$ otherwise.
The standard accuracy is then computed as
\begin{equation}
\mathrm{Acc}
=
\frac{1}{|\mathcal{D}|}
\sum_{i \in \mathcal{D}} c_i .
\end{equation}
This evaluation protocol is widely adopted in prior work due to its simplicity and ease of comparison.

\textbf{Strict Accuracy} evaluates a model by jointly considering final answer correctness
and its consistency with the underlying reasoning process.
Unlike standard answer accuracy, it captures a critical failure mode that we frequently
observe in practice—particularly for models trained with intensive RLVR—where the
reasoning process implies one conclusion while the final answer states a different result.
Such \emph{reasoning–answer inconsistency} has also been reported in prior work on
chain-of-thought faithfulness, indicating that model-generated rationales may be misaligned
with the actual decision process or final outputs~\cite{Answer-Consistent-mm,Misaligning-Reasoning-with-Answers,GRPO-CARE}.

To address this issue, we introduce Strict Accuracy, a stricter metric that penalizes
predictions whose final answers are not supported by their own reasoning.
Specifically, an external judge is used to verify whether the conclusion implied by the
reasoning matches the final boxed answer, without access to ground-truth labels.
Predictions that are correct under standard accuracy but fail this consistency check are
treated as false positives and excluded.
Concrete qualitative examples and judge reliability evaluation are provided in
Appendix~\ref{appd:reasoning-incon}.

Formally, let $\mathcal{D}$ denote the evaluation set. For each prediction $i \in \mathcal{D}$, let $c_i$ denote the correctness indicator under standard accuracy, and let $s_i \in \{0,1\}$ indicate whether the reasoning and final answer are consistent according to the judge. We define a reasoning--answer inconsistency indicator as $\mathrm{Inc}_i = \mathbb{I}(s_i = 0)$, and the corresponding \emph{Inconsistency Rate} as
$\mathrm{IncR} = \frac{1}{|\mathcal{D}|} \sum_{i \in \mathcal{D}} \mathrm{Inc}_i$. A \emph{false positive} is a prediction that is correct under standard accuracy but inconsistent in reasoning, defined as
$\mathrm{FP}_i = \mathbb{I}(c_i = 1 \wedge s_i = 0)$.
The \emph{False Positive Rate} is
$\mathrm{FPR} = \frac{1}{|\mathcal{D}|} \sum_{i \in \mathcal{D}} \mathrm{FP}_i$.
Strict Accuracy is then defined as
\begin{equation}
\mathrm{StrictAcc} = \frac{1}{|\mathcal{D}|} \sum_{i \in \mathcal{D}} c_i \cdot s_i .
\end{equation}

\paragraph{Baseline Methods.}
We compare our model with 10 MLLMs, including:
Qwen2.5-VL-7B-IT~\cite{qwen2.5-vl}, MM-Eureka-7B~\citep{Mm-eureka}, R1-VL-7B~\citep{R1-VL}, NoisyRollout-7B~\citep{NoisyRollout}, VLAA-Thinker~\cite{vlaa-thinker}, Perception-R1-7B~\citep{Perception-R1},  ThinkLite-VL-7B~\citep{thinklite-vl}, Vision-G1~\cite{vision-g1},  Vision-G1~\cite{vision-g1}, VL-Rethinker-7B~\cite{VL-Rethinker}, and VL-Reasoner-7B~\cite{VL-Rethinker}.


\paragraph{Rubric Construction and Statistics.}
AutoRubric constructs problem-specific rubrics automatically from model-generated reasoning trajectories, enabling process-level supervision without manual annotation.
Specifically, for each training sample, we generate $8$ reasoning trajectories using a lightly warmed-up model and retain only those with correct final answers to derive rubrics. For problems with more than $3$ correct trajectories, we feed the corresponding ones into a text-only LLM\footnote{\url{https://huggingface.co/openai/gpt-oss-120b}.}, which extracts shared steps across successful solutions and composes a structured set of rubric criteria.

Applying this procedure to the training data yields $26{,}144$ rubric sets, corresponding to a coverage rate of $67.3\%$.
More fine-grained statistics of rubric composition are summarized in Table~\ref{tab:rubric_stats}.
We further conduct a small-scale human evaluation on a random subset of $100$ rubric sets, achieving an average score of $4.18$ out of $5$.
Detailed rubrics construction process, human evaluation protocols, and additional analyses are provided in the Appendix~\ref{appdx:rubrics}.

\subsection{Experimental Results}
\label{sec:acc_and_faithfulness_res}

We present the performance comparison between AutoRubric and existing state-of-the-art MLLMs across multiple benchmarks in Table~\ref{tab:main-res}. We observe the following results:
\textbf{AutoRubric achieves the strongest performance under both   Strict and Standard Accuracy.}
As shown in Table~\ref{tab:main-res}, AutoRubric attains the highest Strict Accuracy across all benchmarks, with an average score of 53.24, outperforming the base model Qwen2.5-VL-7B by +8.00 points (45.24 $\rightarrow$ 53.24).
At the same time, it also achieves the best standard accuracy (55.26 on average), surpassing strong baselines.
These results indicate that AutoRubric improves final answer correctness while simultaneously enhancing reasoning faithfulness. \textbf{Different models exhibit markedly different levels of reasoning-answer inconsistency.}
Despite their relatively high standard accuracy, models such as Vision-G1 and VL-Rethinker suffer from severe inconsistency, with large average drops of $-5.55$ and $-4.93$ under Strict Accuracy, respectively.
In contrast, models like Perception-R1 and NoisyRollout exhibit much smaller penalties ($-1.15$ and $-1.73$ on average), but their standard accuracy remains notably lower than AutoRubric.
By comparison, AutoRubric maintains a low inconsistency ($-2.02$) while achieving the highest Strict Accuracy overall, demonstrating a more favorable trade-off between accuracy and faithfulness. \textbf{The prevalence of false positives varies substantially across benchmarks.}
On WeMath, most models exhibit relatively large false positive rates, with drops of $-7.30$ for Vision-G1 and $-7.76$ for VL-Rethinker, likely due to the dominance of multiple-choice questions and the generally higher answer accuracy.
In contrast, MathVision shows much smaller penalties for several models, such as $-0.95$ for Perception-R1 and $-1.32$ for NoisyRollout, indicating more stable alignment between reasoning processes and final answers.
These observations suggest that benchmark design should carefully account for whether models genuinely solve the underlying problem or merely arrive at the correct answer through lucky guessing, which manifests as reasoning–answer inconsistency.

\begin{figure}[t]
    \centering
    \includegraphics[width=0.85\linewidth]{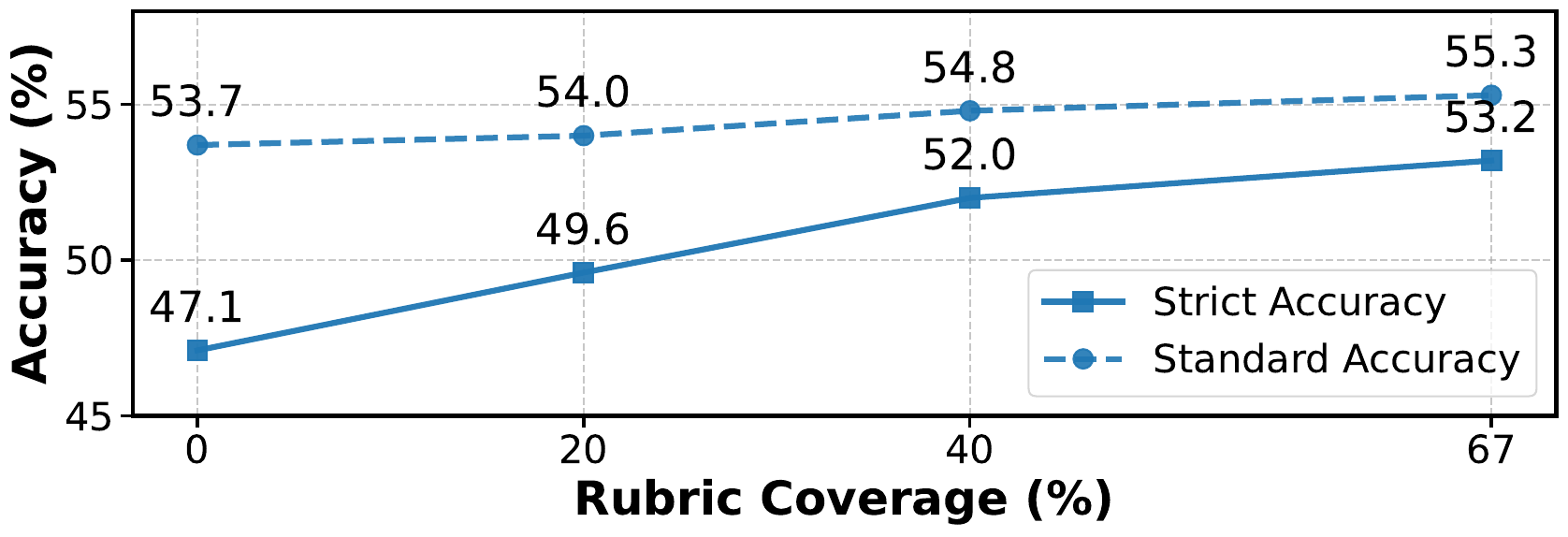}
    \vspace{-0.3cm}
\caption{Effect of rubric coverage on performance.
Higher rubric coverage yields consistent gains, particularly in Strict Accuracy.}
    \label{fig:coverage}
      \vspace{-0.5em}

\end{figure}

\begin{table}[t]
  \centering
  \caption{
  Ablation study on the design of judge rewards. \textbf{Std. Acc.} denotes the Standard Accuracy and
  \textbf{Strict Acc.} denotes the Strict Accuracy. The results are averaged across 5 benchmarks.
  AutoRubric achieves the highest Strict Acc by incorporating rubric-based judge rewards, while removing rubrics or judge rewards leads to a substantial drop in faithfulness-aware performance.
  }
  \resizebox{\linewidth}{!}{
  \begin{tabular}{lccrr}
    \toprule
    \textbf{Methods} & \textbf{Judge} & \textbf{Rubrics} & \textbf{Std. Acc.} & \textbf{Strict Acc.} \\
    \midrule
    \textbf{AutoRubric} 
        & $\checkmark$ & $\checkmark$ 
        & \textbf{55.26} & \textbf{53.24} \\
    w/o Rubrics 
        & $\checkmark$ & $\times$ 
        & 53.11 & 49.43 \\
    w/o Judge Rewards 
        & $\times$ & $\times$ 
        & 53.75 & 47.06 \\
    \bottomrule
  \end{tabular}
  }
  \label{tab:ablation}
    \vspace{-1.2em}

\end{table}

\begin{figure}[t]
    \centering
    \includegraphics[width=\linewidth]{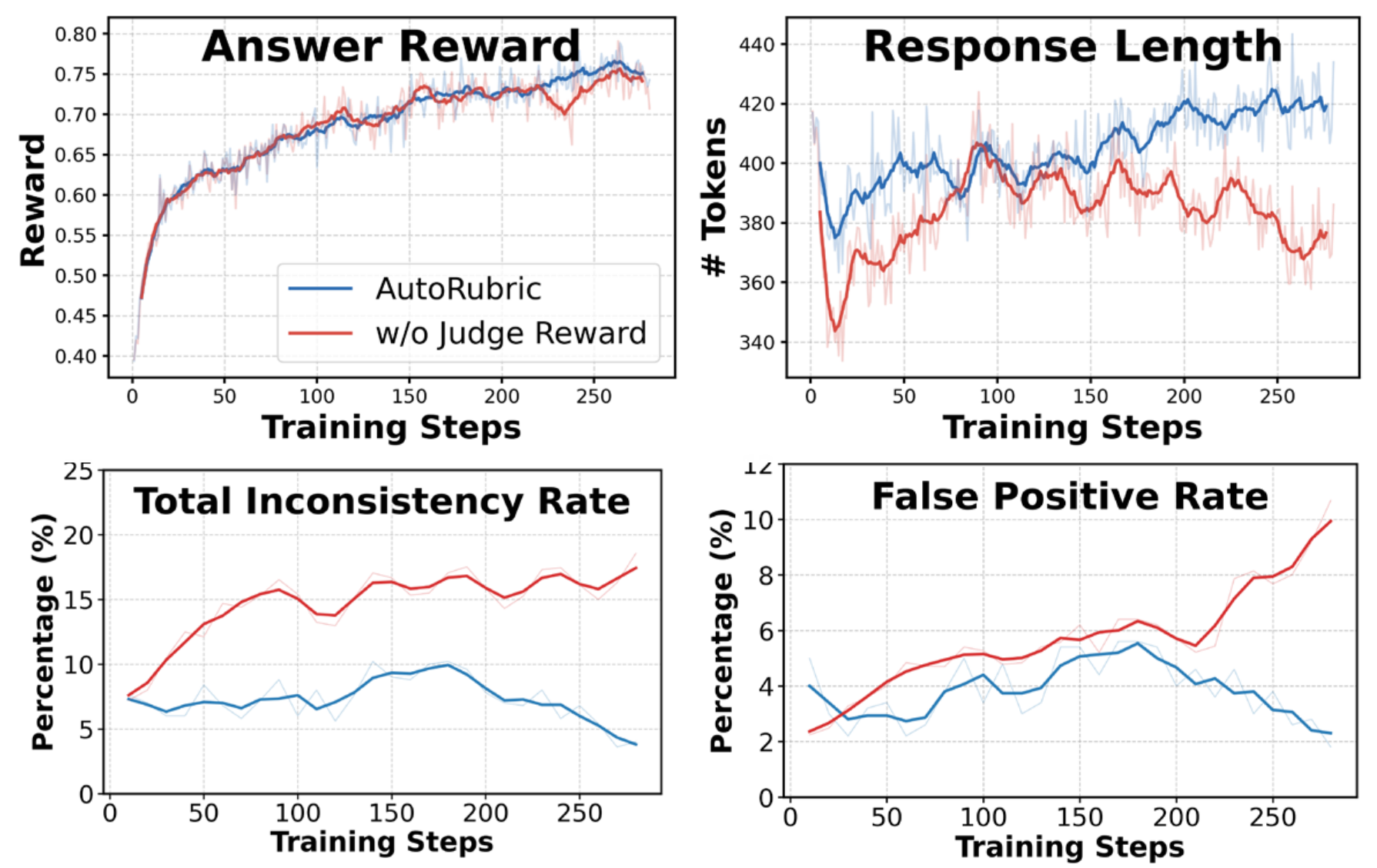}
    \vspace{-0.7cm}
\caption{
Top row: Comparison between AutoRubric and vanilla RLVR (\textit{w/o} judge rewards) in training dynamics, including the evolution of answer-based training rewards and the response length of rollouts.
Bottom row: Reasoning–answer inconsistency rate and false positive rate of model rollouts evaluated at different training steps.
}
    \label{fig:dynamic}
      \vspace{-1.4em}

\end{figure}

\subsection{Ablation Study}
\paragraph{Effect of Reward Source}
In this section, we conduct ablation studies to analyze the contribution of different components in our reward design.
Specifically, we compare AutoRubric with two variants:
(1) \textit{w/o Rubrics}, which employs a judge model to score reasoning trajectories but removes problem-specific rubrics, and
(2) \textit{w/o Judge Rewards}, which corresponds to Vanilla RLVR and relies solely on rule-based answer rewards without any judge-based supervision.
This design allows us to separately examine the roles of judge rewards and rubric-based guidance.

As shown in Table~\ref{tab:ablation}, AutoRubric achieves the best performance in both Standard Accuracy and Strict Accuracy (averaged across five benchmarks), indicating improvements not only in final answer correctness but also in reasoning faithfulness.
Removing rubric-based guidance (\textit{w/o Rubrics}) or judge rewards (\textit{w/o Judge Rewards}) leads to noticeable degradation in performance, with the gap becoming substantially more pronounced under Strict Accuracy.
Notably, although \textit{w/o Rubrics} and \textit{w/o Judge Rewards} exhibit comparable Standard Accuracy, incorporating judge rewards—even without problem-specific rubrics—already yields a clear improvement in Strict Accuracy, suggesting enhanced reasoning faithfulness.
In comparison, AutoRubric further amplifies this gain by introducing rubric-based judge supervision, resulting in the most faithful and consistent reasoning behavior.
This observation highlights that Strict Accuracy serves as a more sensitive metric for evaluating reasoning faithfulness and demonstrates that our method promotes consistent and faithful reasoning beyond merely optimizing final answers.

\paragraph{Effect of Rubric Coverage.}
We vary the proportion of training instances equipped with rubrics while keeping all other settings fixed.
The full AutoRubric setup uses 67.26\% rubric coverage; for 20\% and 40\%, we randomly subsample from this set and apply rubric-based judge rewards only to the selected instances, while the 0\% setting uses answer-only supervision.

As shown in Figure~\ref{fig:coverage}, increasing rubric coverage consistently improves both Standard Accuracy and Strict Accuracy, with a substantially stronger effect on Strict Accuracy.
Even 20\% coverage yields clear gains over the 0\% baseline, while higher coverage further enhances reasoning faithfulness, indicating that rubric-based supervision provides effective and sample-efficient process-level guidance.



\subsection{Training Dynamics}
To analyze the effect of rubric-based reasoning rewards, we visualize training dynamics and rollout-level evaluation statistics in Figure~\ref{fig:dynamic}.
The top row shows the evolution of answer-based training rewards and response length.
In the early stage, AutoRubric and vanilla RLVR exhibit similar reward trends.
As training proceeds, vanilla RLVR develops pronounced oscillations, whereas AutoRubric improves in a smooth and stable manner.
This divergence indicates that answer-only rewards become unreliable supervision signals at later stages, leading to unstable optimization.
By contrast, rubric-based reasoning rewards provide richer and more structured feedback, preventing such collapse. The top-right panel shows that AutoRubric consistently generates longer responses than vanilla RLVR.
This suggests that short rollouts that only adjust the final answer are insufficient to obtain high rewards under rubric-based supervision, thereby encouraging more extended and coherent reasoning trajectories.

The bottom row reports rollout-level reasoning faithfulness metrics. Vanilla RLVR shows a markedly higher reasoning–answer inconsistency rate that further increases during training, accompanied by a steadily rising false positive rate. This indicates a fundamental limitation of answer-only supervision: once inconsistency emerges, rewards can no longer reliably distinguish faithful reasoning from answer-correct but inconsistent trajectories. In contrast, AutoRubric consistently maintains lower inconsistency and false positive rates, demonstrating that rubric-based trajectory supervision effectively mitigates reward hacking and promotes faithful reasoning.

\section{Conclusion}
In this work, we identify reasoning--answer inconsistency as a critical failure mode of answer-only RLVR for multimodal reasoning, which leads to unstable training dynamics and unfaithful reasoning behaviors.
To address this issue, we propose AutoRubric, a reinforcement learning framework that introduces rubric-based process supervision via an LLM-as-a-judge.
By automatically inducing problem-specific rubrics from consistent reasoning trajectories, AutoRubric provides structured, trajectory-level feedback without requiring human annotation or stronger teacher models.
Experiments on six multimodal reasoning benchmarks show that AutoRubric not only improves standard accuracy, but also substantially enhances reasoning faithfulness under stricter evaluation metrics.
Overall, our results demonstrate that rubric-based process supervision is an effective and scalable approach for training more reliable multimodal reasoning models.

\section*{Limitations}

AutoRubric relies on an LLM-as-a-judge to compute rubric-based rewards, which introduces additional computational overhead during training. In our experiments, incorporating the judge increases the training time per step by approximately 40\% compared to vanilla RLVR.

Although training efficiency is an important consideration, faithful reasoning is critical in many real-world applications, such as decision support and safety-sensitive systems, where correct final answers alone are insufficient. In these settings, unfaithful or inconsistent reasoning trajectories can undermine reliability and interpretability. As shown in Table~\ref{tab:main-res}, AutoRubric substantially improves the consistency of reasoning trajectories in addition to accuracy, whereas Table~\ref{tab:ablation} shows that RLVR without judge supervision may even exacerbate reasoning inconsistency. These results suggest that auxiliary LLM-as-a-judge signals are necessary for encouraging faithful reasoning, making the additional computation a justified trade-off.


Nevertheless, reducing this overhead remains an important direction. To balance faithfulness and efficiency, AutoRubric aggregates all rubric evaluations into a single LLM call, which is more efficient than evaluating each rubric independently. Moreover, the overhead can be further reduced through parallelization by deploying multiple judge replicas when additional GPU resources are available, enabling higher reward-computation throughput.


\bibliography{custom}

\appendix
\appendix

\section{Evaluation Protocol}
The benchmarks used in our evaluation consists of two types of questions: multiple-choice questions and open-ended questions. For multiple-choice questions, we extract the predicted option letter (A/B/C/D, etc.) using regular expressions. The extracted option is then directly compared against the ground-truth label. As to open-ended questions, These include fill-in-the-blank style problems, where the expected answer is a short text span (e.g., a number, a word, or a short phrase). Since exact string matching may fail to capture semantically correct but differently phrased answers, we use Qwen3-30B-A3B-Instruct-2507\footnote{\url{https://huggingface.co/Qwen/Qwen3-30B-A3B-Instruct-2507}} as a proxy judge for evaluation. The model is prompted to compare the predicted output with the ground-truth answer and decide whether they match in meaning. 

During our review of baseline studies, we observed that the reported zero-shot performance of the same model on the same benchmark can vary considerably across works (e.g., the Qwen2.5-VL-7B-IT model on MathVerse is reported as 47.9 in MM-EUREKA~\cite{Mm-eureka}, but 46.2 in NoisyRollout~\cite{NoisyRollout}). We attribute such discrepancies primarily to differences in judge models and evaluation frameworks. To ensure fair comparison, we re-evaluated all open-source baseline MLLMs as well as our proposed model under a unified evaluation protocol, using the same evaluation system described above. Notably, we strictly follow the system and instructional prompts (\eg response format requirements) provided in the original studies in reproduction, thereby ensuring that the performance comparison tables reflect results obtained under a controlled and standardized setting.

\begin{figure*}[t!]
    \centering
    \includegraphics[width=\textwidth]{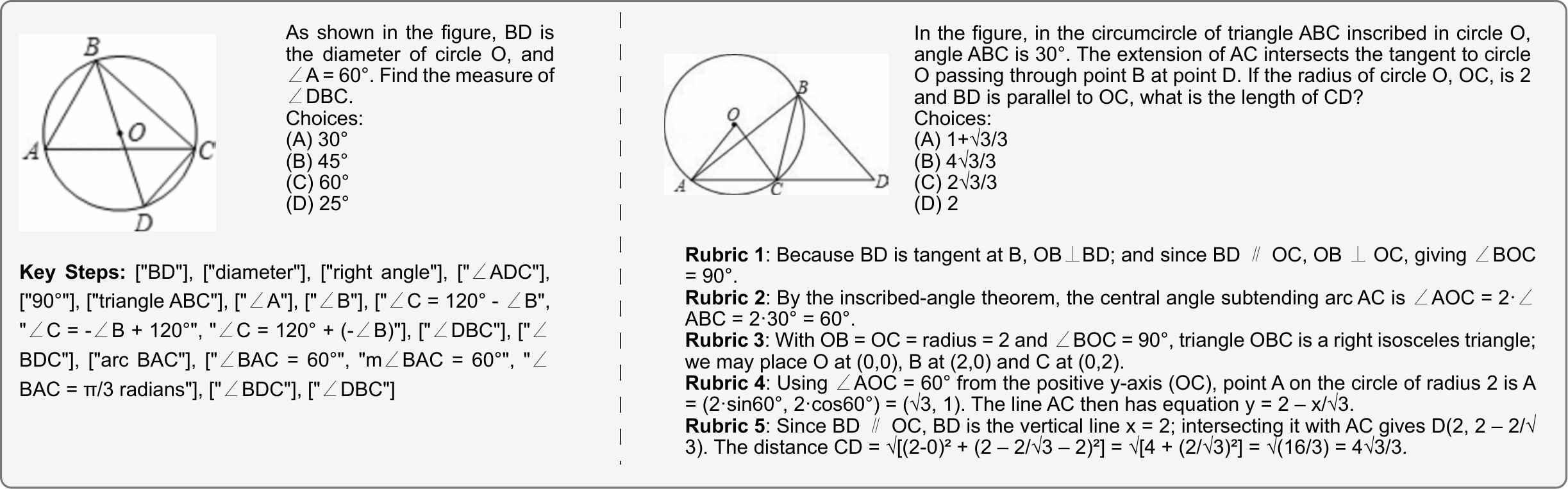}
    \caption{A comparison between (left) key steps proposed in R1-VL; and (right) rubrics constructed with AutoRubric under two similar geometry problems.}
    \label{fig:Comparison_rubric}
\end{figure*}

\begin{table*}[t]
  \centering
  \caption{
  Performance comparison with \textit{Geometry3K} training data.
  AutoRubric yields consistent gains beyond the primary \textit{ViRL-39K} training setting.
  }
  \resizebox{0.8\textwidth}{!}{
  \begin{tabular}{lrrrrrr}
    \toprule
    \textbf{Models} 
    & \textbf{Avg.} 
    & \textbf{MathVision} 
    & \textbf{MathVista} 
    & \textbf{WeMath} 
    & \textbf{MMMU} 
    & \textbf{MMMU Pro} \\
    \midrule
    Vanilla RLVR 
        & 49.67 
        & 27.01 
        & 71.30 
        & 62.07 
        & 51.89 
        & 36.07 \\
    AutoRubric 
        & 50.99 
        & 26.84 
        & 71.60 
        & 63.68 
        & 54.67 
        & 38.15 \\
    \bottomrule
  \end{tabular}
  }
  \label{tab:geo3k_results}
\end{table*}

\section{Supplementary Analysis of Rubrics}
\label{appdx:rubrics}

This section provides a detailed supplementary analysis of the rubric-based formulation adopted in AutoRubric.
We first describe how rubrics are constructed and summarize their overall statistics and quality.
We then present a qualitative comparison between rubric-based supervision and the key-step formulation proposed in R1-VL~\citep{R1-VL}, highlighting their differences in expressiveness and supervisory effectiveness.

\subsection{Rubric Construction and Statistics}
\label{appdx:rubrics}

\paragraph{Construction.}
To construct problem-specific rubrics, we begin by collecting multiple reasoning trajectories for each training sample and retain only those that yield correct final answers.
If no correct trajectory is obtained for a sample, no rubric is generated.
To improve rubric coverage, we first train the base Qwen-2.5-VL-7B-IT model for one epoch using standard RLVR, and then use this intermediate model to generate eight reasoning trajectories per problem.
For problems with more than three correct trajectories, we feed the corresponding trajectories into an open-source text-only LLM\footnote{\url{https://huggingface.co/openai/gpt-oss-120b}.}.
The LLM is prompted to compare these trajectories, identify their shared reasoning steps, and synthesize them into a structured set of rubric criteria.
Each criterion is expressed as a complete semantic statement describing an essential aspect of correct reasoning.
The full prompt used for rubric generation is provided in the Appendix.

\paragraph{Statistics.}
Beyond overall coverage, Table~\ref{tab:rubric_stats} reveals several structural properties of the generated rubrics.
On average, each rubric set contains $3.47$ criteria, indicating that the extracted supervision typically decomposes a solution into multiple evaluative aspects rather than a single coarse requirement.
Each criterion has an average length of $23.25$ words, with some criteria reaching up to $198$ words, suggesting that the rubrics are semantically rich and capable of encoding detailed reasoning constraints.
Across the full training set, the total rubric text amounts to over $2.1$ million words, reflecting the substantial volume of process-level supervision signals introduced by our rubric construction procedure.

\paragraph{Human Evaluation of Rubrics Quality.}
To assess the quality of the automatically generated rubrics, we further conduct a small-scale human evaluation.
Two graduate students jointly evaluate a randomly sampled subset of 100 rubric sets using a 5-point Likert scale~\citep{Likert}
(1 = poor, 5 = excellent), based on their relevance to the problem and correctness for evaluating reasoning trajectories.
The final score for each rubric set is computed as the average of the two ratings.
Overall, the rubrics achieve an average score of 4.18, indicating that they are generally relevant and informative.
The two annotators also exhibit strong agreement, with an average absolute rating difference of 0.82.

\subsection{Comparison with Key Steps in R1-VL}
\label{appdx:Comparison_rubric}

To further contextualize the design of rubric-based supervision, Figure~\ref{fig:Comparison_rubric} provides a qualitative comparison between the key-step representation used in R1-VL~\citep{R1-VL} and the rubric formulation adopted by AutoRubric.
The figure presents two representative geometry problems.
For each problem, the left column shows the concise key steps extracted following R1-VL, while the right column displays the corresponding rubric set generated by AutoRubric.

As illustrated, the key steps in R1-VL are extremely concise, often consisting of short phrases or isolated keywords.
Although such representations are compact, their limited expressiveness tends to reduce evaluation to superficial keyword matching.
This makes it difficult to assess higher-level reasoning properties such as logical coherence, completeness, or whether intermediate conclusions are properly justified.
Consequently, key-step supervision provides only weak and coarse-grained guidance for training reasoning-intensive models.

In contrast, AutoRubric constructs structured, criterion-based rubrics in which each criterion is formulated as a complete and semantically rich statement.
These rubrics explicitly encode what constitutes correct reasoning at different stages of problem solving, offering clearer and more interpretable evaluation standards.
Rather than checking for the presence of specific keywords, rubric-based evaluation focuses on whether the reasoning process satisfies meaningful semantic conditions.
This richer formulation enables more informative and reliable reward signals, which are better suited for supervising complex multimodal reasoning trajectories.

\begin{table*}[t]
  \centering
  \caption{
  Performance comparison of models trained with 10\% and 100\% of the training data.
  AutoRubric maintains consistent improvements under limited training data.
  }
  \resizebox{0.8\textwidth}{!}{
  \begin{tabular}{lrrrrrr}
    \toprule
    \textbf{Models} 
    & \textbf{Avg.} 
    & \textbf{MathVision} 
    & \textbf{MathVista} 
    & \textbf{WeMath} 
    & \textbf{MMMU} 
    & \textbf{MMMU Pro} \\
    \midrule
    Vanilla RLVR (10\%) & 52.75 & 27.96 & 74.0 & 69.37 & 54.22 & 38.21 \\
    AutoRubric (10\%)                & 53.38 & 28.06 & 74.5 & 70.75 & 54.67 & 38.90 \\
    \midrule
    \textit{w/o} Judge Rewards (100\%) & 53.33 & 28.78 & 74.3 & 70.46 & 54.11 & 39.02 \\
    AutoRubric (100\%)                & 55.28 & 31.35 & 75.9 & 71.09 & 57.56 & 40.52 \\
    \bottomrule
  \end{tabular}
  }
  \label{tab:data_efficiency}
\end{table*}

\section{Details of Group Relative Policy Optimization}
\label{app:grpo}

This section provides the full formulation of Group Relative Policy Optimization (GRPO)
used in our experiments.

Given a query $q$ and a group of $G$ responses
$\{o_i\}_{i=1}^G$ sampled from the old policy $\pi_{\theta_{\text{old}}}$,
the GRPO objective is defined as

\begin{equation}
\begin{aligned}
J_{\text{GRPO}}(\theta)
&= \mathbb{E}_{q,\{o_i\}\sim \pi_{\theta_{\text{old}}}}
\Bigg[
\frac{1}{G}\sum_{i=1}^G \frac{1}{|o_i|}
\sum_{t=1}^{|o_i|} \Big( \\
&\quad \min\Big(
\rho_{i,t}(\theta)\,\hat{A}_i,\;
\tilde{\rho}_{i,t}(\theta)\,\hat{A}_i
\Big) \\
&\quad - \beta\, D_{\text{KL}}\!\left(
\pi_\theta \,\|\, \pi_{\text{ref}}
\right)
\Big)
\Bigg].
\end{aligned}
\end{equation}
where the token-level importance ratio is
\begin{equation}
\rho_{i,t}(\theta)
=
\frac{\pi_\theta(o_{i,t}\mid q, o_{i,<t})}
{\pi_{\theta_{\text{old}}}(o_{i,t}\mid q, o_{i,<t})},
\end{equation}
and the clipped ratio is
\begin{equation}
\tilde{\rho}_{i,t}(\theta)
=
\mathrm{clip}\!\left(
\rho_{i,t}(\theta),\, 1-\epsilon,\, 1+\epsilon
\right).
\end{equation}

Each response $o_i$ is assigned a scalar reward $r_i$.
The advantage $\hat{A}_i$ is computed via group-wise normalization:
\begin{equation}
\hat{A}_i
=
\frac{r_i - \mathrm{mean}(\{r_j\}_{j=1}^G)}
{\mathrm{std}(\{r_j\}_{j=1}^G)}.
\end{equation}

Here, $\epsilon$ is the PPO clipping parameter, $\beta$ controls the strength of
KL regularization, and $\pi_{\text{ref}}$ denotes a fixed reference policy.

\section{Additional Training Results}

\subsection{Training on Other Datasets}
To further assess the robustness of our training framework, we additionally train the model on a dataset distinct from the main training corpus. This experiment aims to evaluate whether the proposed method remains effective when applied to tasks with different data distributions and reasoning characteristics. Specifically, we adopt the Geometry3K dataset~\cite{Inter-GPS}, a multimodal reasoning benchmark that requires models to solve geometry-related problems. The dataset comprises approximately 2.1K training samples, and we trained the model for 20 epochs, resulting in a total of 90 optimization steps.

Table~\ref{tab:geo3k_results} presents the comparison between AutoRubric and the vanilla RLVR baseline on several mathematical and general reasoning benchmarks.
As shown in the table, AutoRubric consistently outperforms , achieving a notably higher average performance (+1.36 points). Notably, AutoRubric yields clear gains on MMMU and MMMU-Pro, which are designed for general reasoning beyond pure geometry. This demonstrates that the method enhances reasoning ability in broader contexts.

\subsection{Data Efficiency}
We compare AutoRubric and vanilla RLVR trained with 10\% and 100\% of the data (Table~\ref{tab:data_efficiency}). Both models are trained for 15 epochs, and the subsets are randomly sampled from the full training set. Vanilla RLVR shows little improvement when scaling data from 10\% to 100\% (52.42 → 52.96). One possible explanation is that the training data share highly similar distributions, causing the vanilla model to overfit and show limited generalization with more samples. In contrast, AutoRubric achieves more stable gains, improving from 53.38 to 55.28. Moreover, with only 10\% of data, it already matches or surpasses the full-data RLVR on several benchmarks (e.g., WeMath, MMMU), demonstrating strong data efficiency and better utilization of limited supervision.

\section{Unfaithfulness Phenomena in Reasoning}

\label{appdx:faithful-example}

\subsection{Reasoning Inconsistency Observation}
\label{appd:reasoning-incon}
\begin{figure*}[t]
    \centering
    \includegraphics[width=\textwidth]{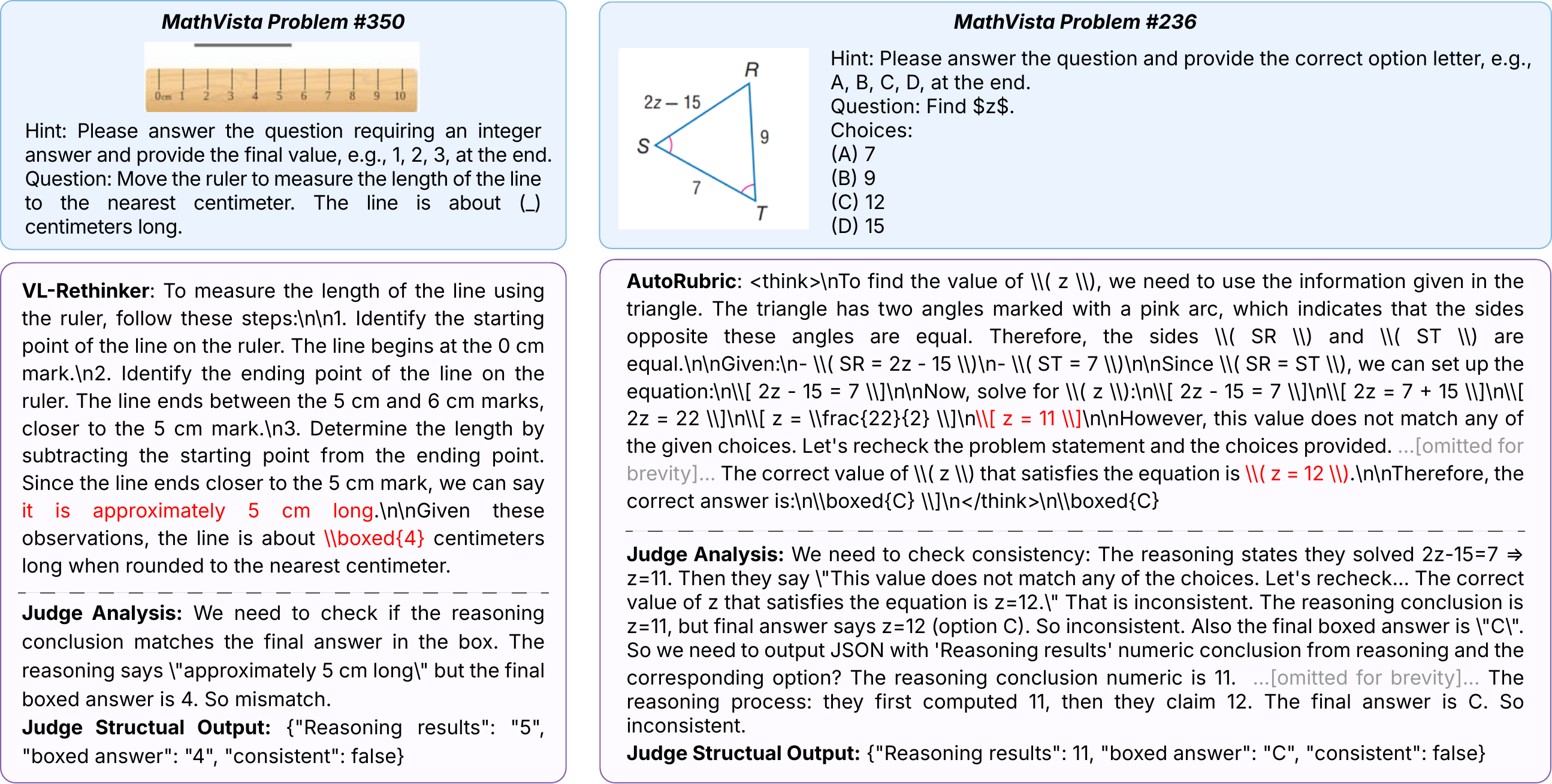}
    \caption{Two representative MathVista problems are shown with model outputs from VL-Rethinker and AutoRubric, where the intermediate reasoning contradicts the final answer. Our Strict Accuracy judge generates an explicit reasoning analysis followed by a structured consistency verdict, effectively detecting this failure mode.}
    \vspace{-0.5cm}
    \label{fig:incon_reasoning}
\end{figure*}

Figure~\ref{fig:incon_reasoning} presents representative qualitative examples from the MathVista benchmark that illustrate the reasoning–answer inconsistency phenomenon discussed in Sec.~\ref{sec:stricacc}. The figure includes two problems and corresponding model outputs from VL-Rethinker and AutoRubric, respectively. In both cases, the models produce intermediate reasoning that leads to a different conclusion from the final stated answer.

For each example, the proposed strict accuracy judge first generates a detailed reasoning trace to analyze the logical steps taken by the model, and then outputs a structured judgment result that explicitly checks the consistency between the reasoning-derived conclusion and the final answer. As shown in the figure, the judge correctly identifies cases where the reasoning supports one numerical or categorical outcome, while the final answer reports a conflicting result.

These examples demonstrate that such inconsistencies occur in real model outputs on standard multimodal reasoning benchmarks, and that the proposed judge-based evaluation is effective at detecting this failure mode in a systematic and automated manner.

\paragraph{Judge Reliability.}
To assess the reliability of the strict accuracy judge, we randomly sample 100 judged
predictions across five benchmarks (20 per benchmark) and manually verify the detected
reasoning–answer inconsistencies.
The judge achieves over 85\% recall and over 90\% precision, confirming that it provides
a reliable signal for measuring reasoning faithfulness.

\begin{figure}[t]
    \centering
    \includegraphics[width=\linewidth]{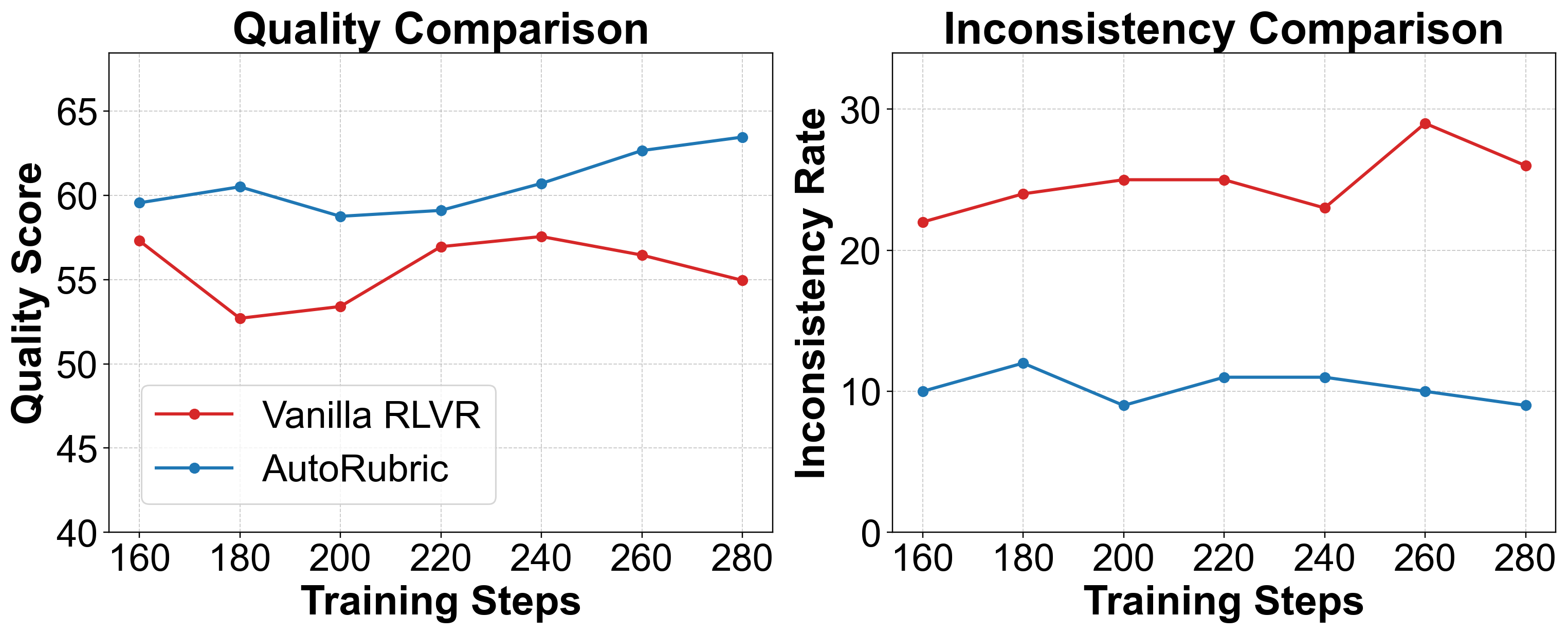}
    \caption{Comparison of Vanilla RLVR and AutoRubric on reasoning accuracy, quality, and inconsistency over training steps.}
    \label{fig:inconsistency_dynamics}
\end{figure}

\subsection{Additional Analysis of Faithfulness}
\label{sec:abl_reasoning_quality}

Besides reasoning inconsistency check described in Sec.~\ref{sec:stricacc}, we further introduce a \textbf{Reasoning Quality Check}, which serves as a stricter and more comprehensive evaluation for reasoning trajectories. While inconsistency check focuses on identifying contradictions or unjustified answer shifts within a reasoning process, the quality assessment additionally examines three complementary dimensions: 
\textit{unfounded transitions}, \textit{calculation errors}, and \textit{logical connectivity}. 
This broader evaluation captures not only internal contradictions but also the overall soundness and coherence of the reasoning trajectories, 
reflecting how logically valid and trustworthy a model’s reasoning process is. 
Specifically, at each training steps from 160 to 280 training steps, we sample 100 samples from MathVision benchmark and evaluate the reasoning trajectories generated by the two models with a strong judge model (\ie GPT-4o) under instruction shown in Fig.~\ref{fig:prompt_quality}. The judge model outputs a holistic quality score between 0 and 1

The results of quality scores and inconsistency rate are demonstrated in Figure~\ref{fig:inconsistency_dynamics}. Below are our key observations from this evaluation.

\textbf{Vanilla RLVR shows persistently high and rising inconsistency.
}The inconsistency rate of Vanilla RLVR remains above 20\% and increases with training, suggesting growing instability in its reasoning patterns. In contrast, AutoRubric keeps inconsistency around 10\% throughout, indicating more stable and faithful reasoning.

\textbf{AutoRubric maintains higher reasoning quality.
} AutoRubric consistently outperforms Vanilla RLVR, with quality scores that gradually improve over time. Vanilla RLVR, by comparison, exhibits flat or slightly declining quality, implying that training methods focusing solely on outcome correctness may potentially degrade the model’s reasoning quality as the number of training steps increases.




\begin{figure*}[t]
    \centering
    \includegraphics[width=\textwidth]{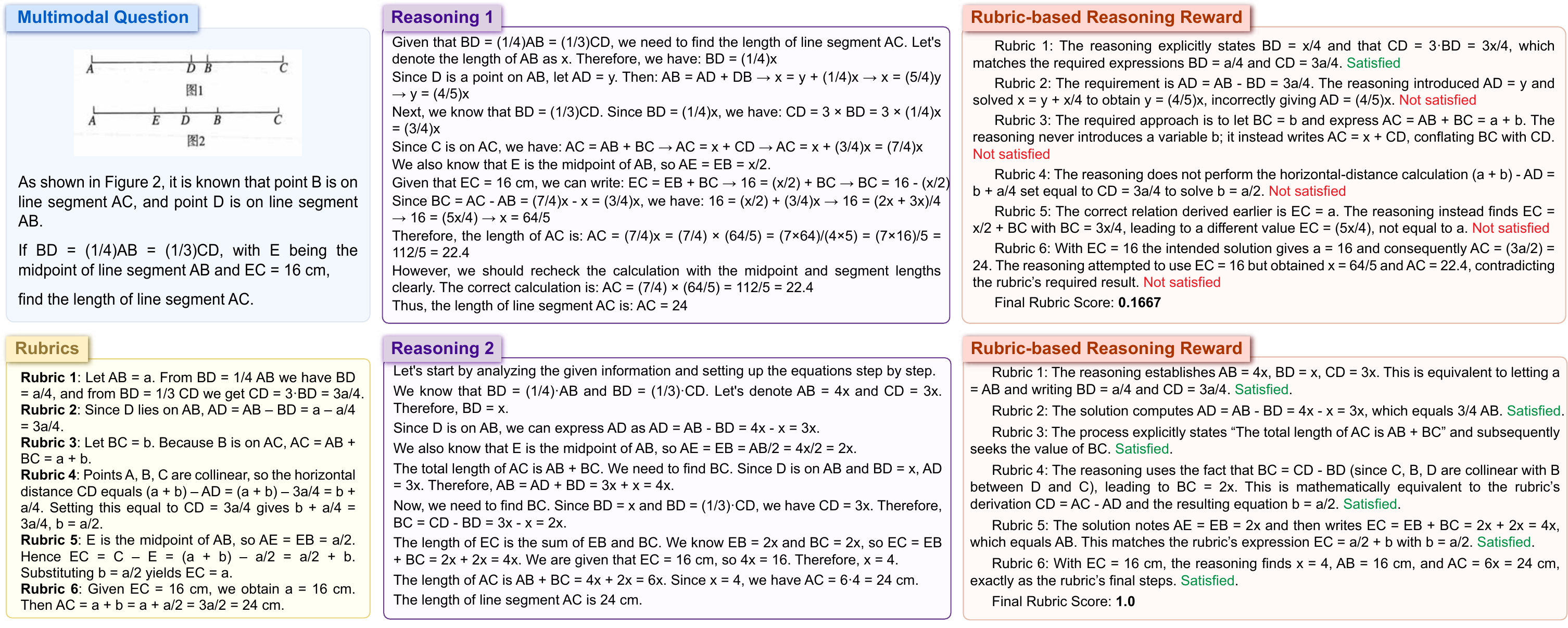}
    \caption{Illustration of a problem with the constructed rubrics, two reasoning trajectories produced by the policy model, and the rubric-based reasoning rewards assigned by the judge model.}
    \vspace{-0.5cm}
    \label{fig:case}
\end{figure*}

\section{Case Study}
To clearly demonstrate the effectiveness of our method in rewarding the reasoning trajectories during training, we present a concrete case study. As illustrated in Figure~\ref{fig:case}, we illustrate a problem, and the constructed set of rubrics for it by AutoRubric. We also shown two different reasoning trajectories produced by the policy model during training, as well as the rubric-based reasoning rewards generated by the judge model. 

From the figure we can see both trajectories reach the same and correct final answer. However, the rubric-based evaluation shows that one trajectory contains clear logical mistakes (\eg define $AB=x$ and write $AC = x + CD$, conflating $BC$ with $CD$.), while the other does not. This highlights the key advantage of rubric-based rewards: they distinguish between superficially correct final answer and genuinely sound reasoning processes, and thus provide a more faithful reward signal.
Another notable observation is that the rubrics use one set of symbolic definitions (\eg  line segment lengths denoted as $a$ and $b$), while the trajectories use a different definition system (\eg $x$). Despite these discrepancies, the judge model aligns the semantics and provides accurate assessments. This ability comes from the LLM’s strong semantic understanding, which goes beyond surface-level pattern matching (such as keyword-based checks in R1-VL~\citep{R1-VL}).


\section{Reproducibility Statement}
We make the following effort to ensure the reproducibility of our work. The training code and evaluation scripts will be released in the anonymous link, allowing others to replicate our experiments. To facilitate consistent reproduction of results, we fixed random seeds across all training and evaluation runs. Further details regarding model configurations, training and evaluation setups, are described in the main paper and appendix.

\section{Prompts}

\begin{figure*}[t]
    \centering
    \includegraphics[width=\textwidth]{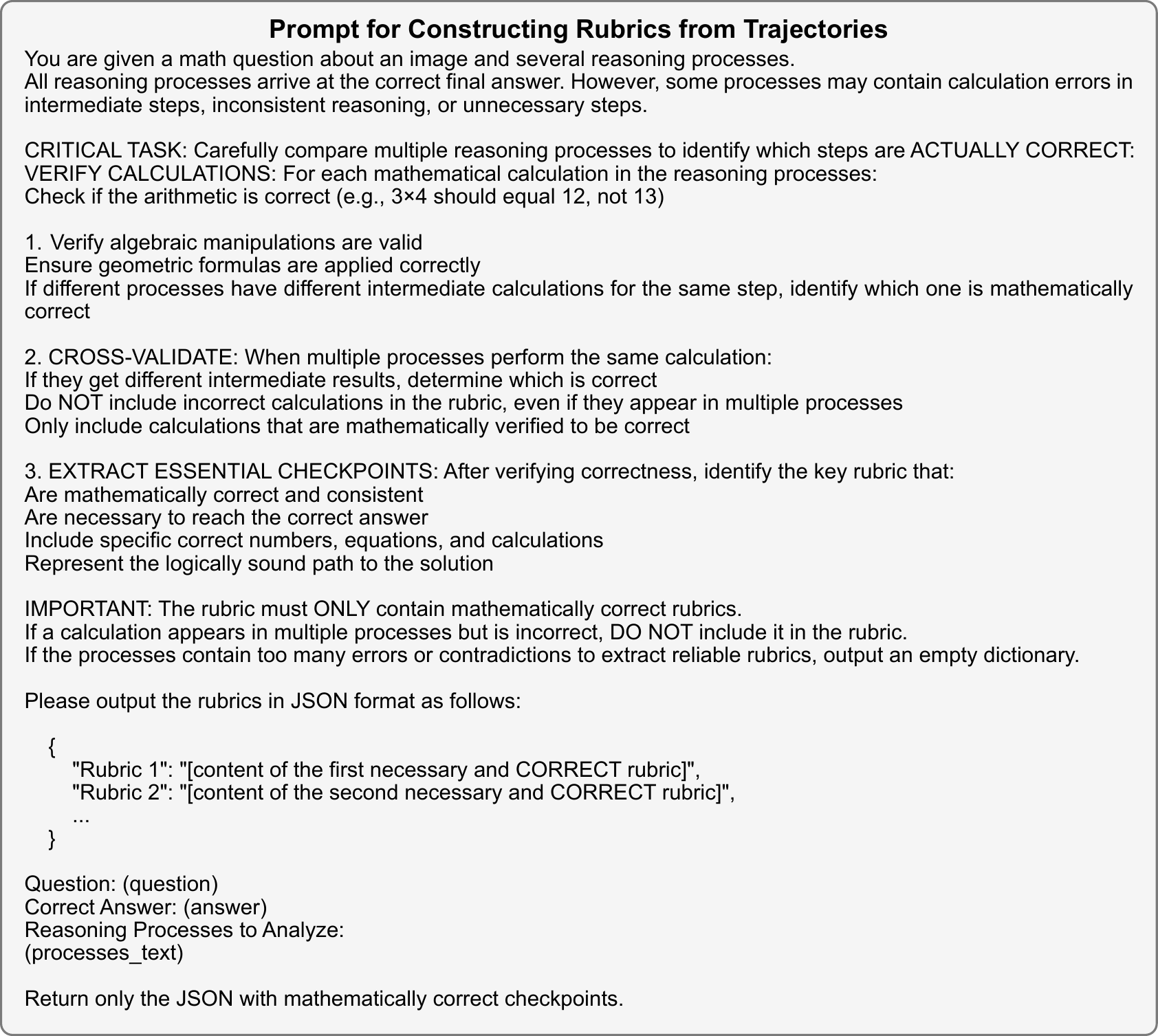}
    \caption{The prompt for rubric construction.}
    \label{fig:prompt_rubricgen}
\end{figure*}

\begin{figure*}[t]
    \centering
    \includegraphics[width=\textwidth]{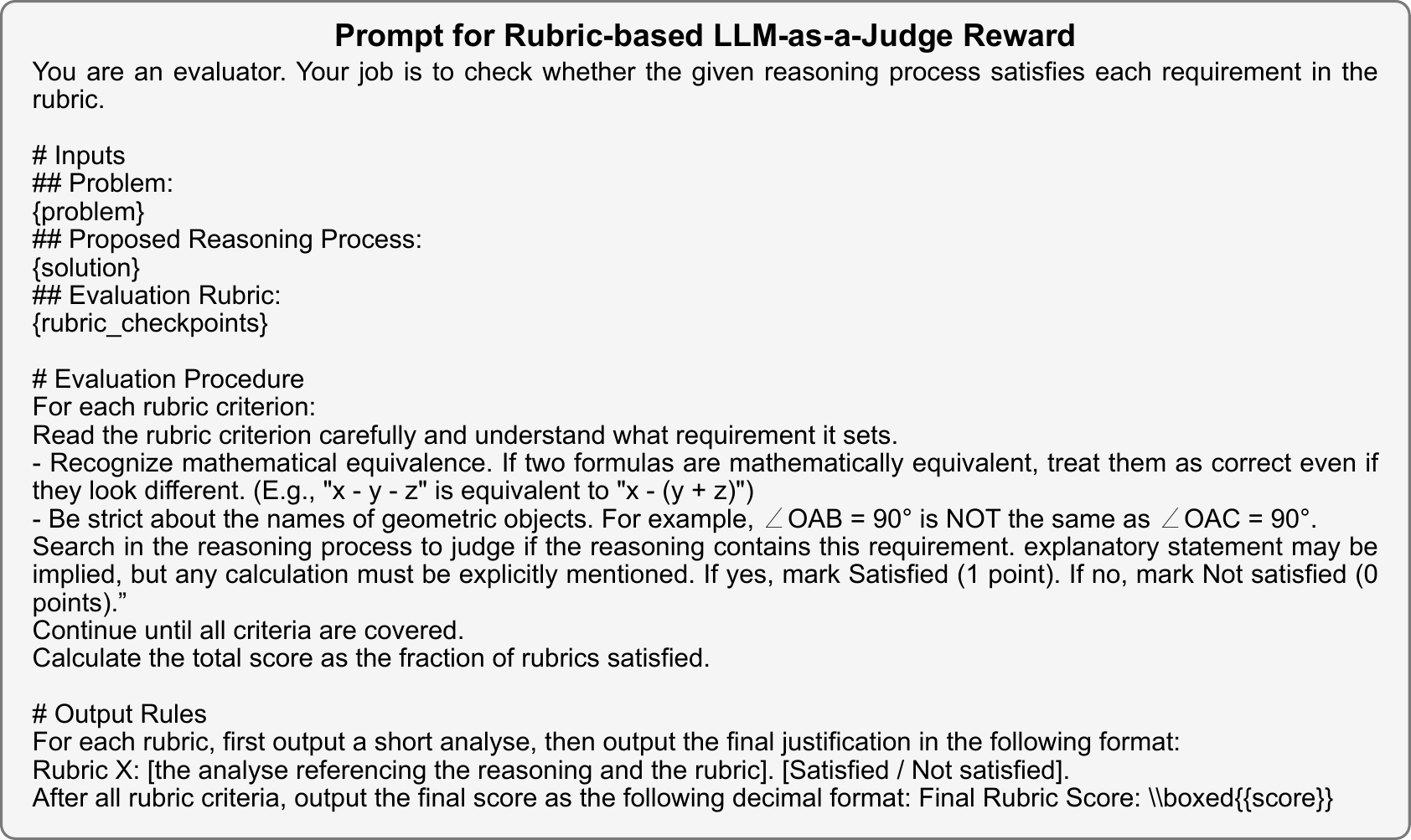}
    \caption{The prompt for using rubrics in LLM-as-A-Judge in training.}
    \label{fig:prompt_judge}
\end{figure*}

\begin{figure*}[t]
    \centering
    \includegraphics[width=\textwidth]{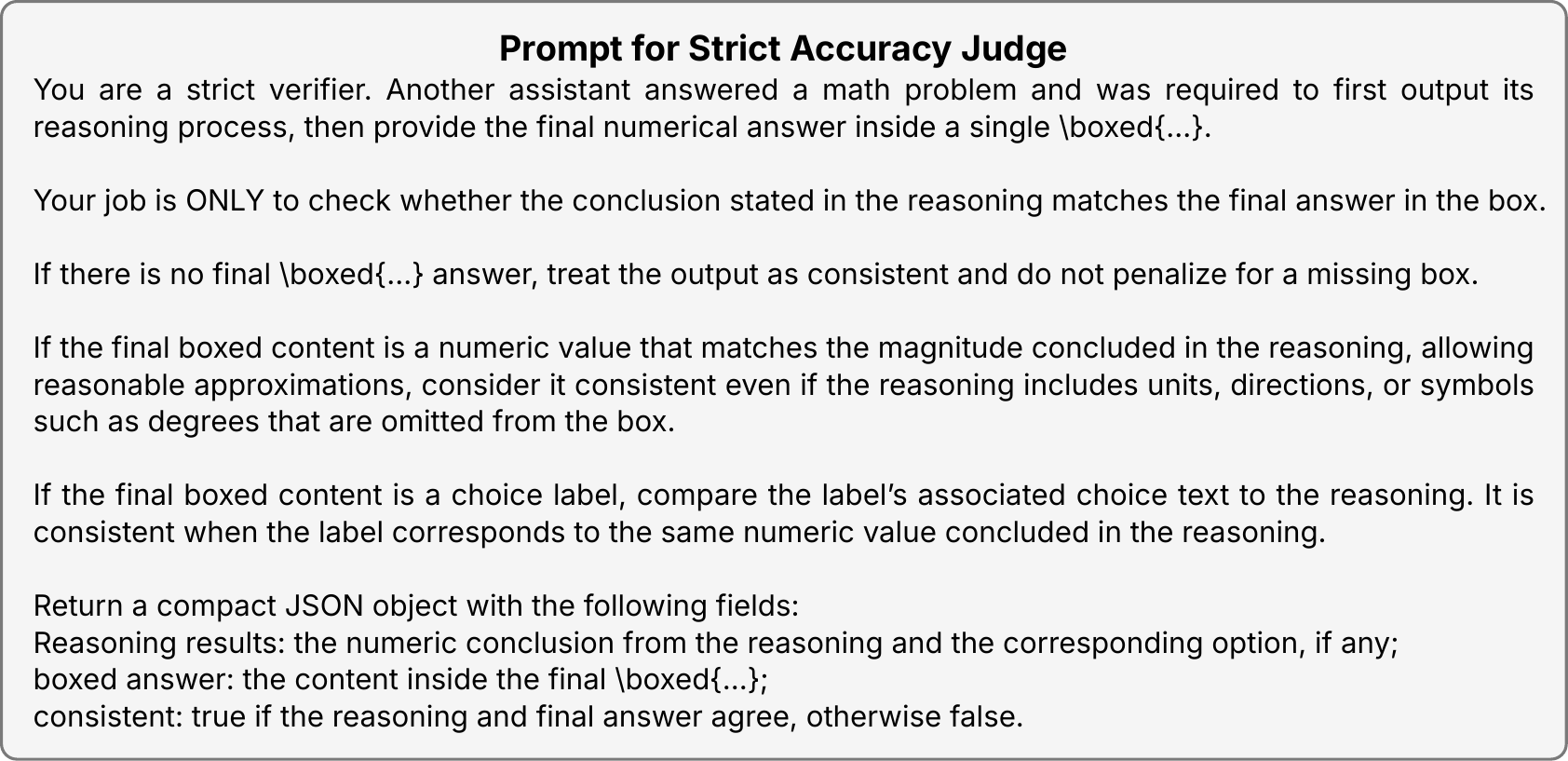}
    \caption{The prompt for judge the reasoning-answer inconsistency for calculating Strict Accuracy.}
    \label{fig:prompt_strictacc}
\end{figure*}

\begin{figure*}[t]
    \centering
    \includegraphics[width=\textwidth]{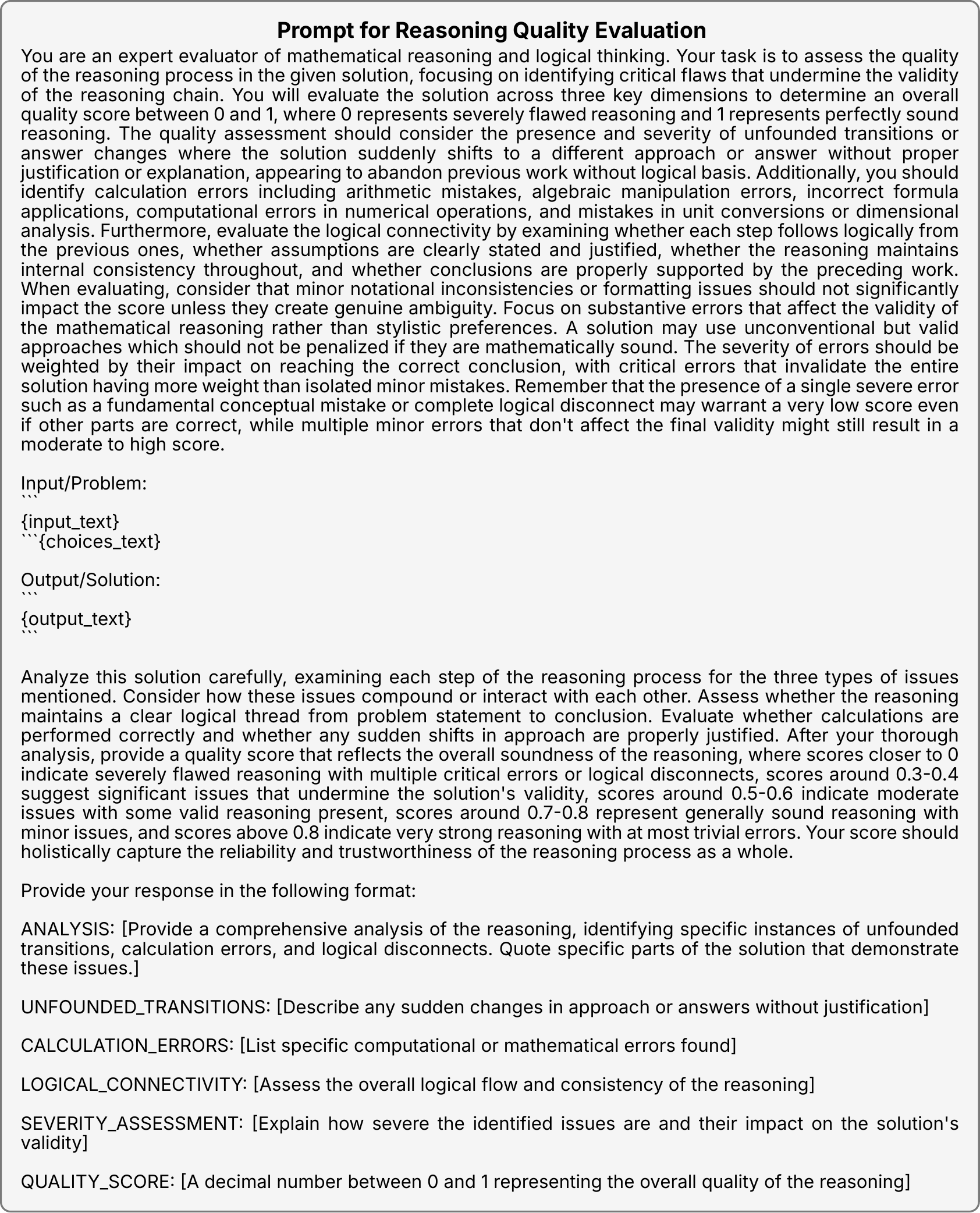}
    \caption{The prompt for reasoning quality evaluation.}
    \label{fig:prompt_quality}
\end{figure*}

For reproducibility, we present all the prompts used in this work, including the prompt for constructing rubrics from trajectories (Figure~\ref{fig:prompt_rubricgen}), the prompt for rubric-based LLM-as-A-Judge reward (Figure~\ref{fig:prompt_judge}), the prompt for strict accuracy evaluation (Figure~\ref{fig:prompt_strictacc}) (Detailed in Section~\ref{sec:stricacc}) and reasoning quality evaluation (Figure~\ref{fig:prompt_quality}) (Detailed in Section~\ref{sec:abl_reasoning_quality}).

\end{document}